\documentclass[accepted]{uai2024}

\usepackage[american]{babel}

\usepackage[authoryear]{natbib}
\usepackage[utf8]{inputenc}
\usepackage[T1]{fontenc}
\usepackage{hyperref}
\usepackage{url}    
\usepackage{booktabs}
\usepackage{amsfonts}       
\usepackage{nicefrac}      
\usepackage{microtype}    
\usepackage{xcolor}    
\usepackage{arydshln}

\usepackage{multirow}
\usepackage{subcaption}
\usepackage{dashbox}
\usepackage{adjustbox}
\usepackage{dsfont}
\newcommand{\1}{\mathds{1}}

\usepackage{tikz}
\usetikzlibrary{positioning, calc, shapes.geometric, shapes.multipart,
  shapes, arrows.meta, arrows,
  decorations.markings, external, trees}
\tikzstyle{line} = [draw, -latex']
\tikzstyle{Arrow} = [
        thick,
        decoration={
                markings,
                mark=at position 1 with {
                        \arrow[thick]{latex}
} },
        shorten >= 3pt, preaction = {decorate}
        ]

\usepackage{amsmath,amssymb,amsthm}
\usepackage{mathtools}

\theoremstyle{definition}
\newtheorem{example}{Example}
\newtheorem{proposition}{Proposition}

\newcommand{\bcd}{\boldsymbol\cdot}
\newcommand{\acronym}{DRIFT}
\newcommand{\longname}{distributional regression using inverse flow transformations}

\newcommand{\calY}{\mathcal{Y}}
\newcommand{\calX}{\mathcal{X}}

\newcommand{\RR}{\mathbb{R}}
\newcommand{\dd}{\mathrm{d}}

\newcommand*{\doi}[1]{\href{http://dx.doi.org/#1}{doi:#1}}

\title{How Inverse Conditional Flows Can Serve as a Substitute for%
\newline Distributional Regression} 

\author[1]{Lucas Kook}
\author[2,3]{Chris Kolb}
\author[2]{Philipp Schiele}
\author[4]{Daniel Dold}
\author[4]{Marcel Arpogaus}
\author[5]{Cornelius Fritz}
\author[6]{Philipp~F.~Baumann}
\author[2,3]{Philipp Kopper}
\author[2,3]{Tobias Pielok}
\author[2,3]{Emilio Dorigatti}
\author[2,3]{\href{mailto:<david@stat.uni-muenchen.de>?Subject=Your UAI 2024 paper}{David R\"ugamer}{}}
\affil[1]{%
   Institute for Statistics and Mathematics\\
   Vienna University of Economics and Business%
}
\affil[2]{%
    \mbox{Department of Statistics\\
    LMU Munich}%
}
\affil[3]{%
    Munich Center for Machine Learning (MCML)%
  }
\affil[4]{%
HTWG Konstanz%
}
\affil[5]{%
Department of Statistics\\
The Pennsylvania State University%
}
\affil[6]{%
KOF Swiss Economic Institute\\
ETH Zurich%
}

\begin{document}

\maketitle

\begin{abstract}
Neural network representations of simple models, such as linear regression, are
being studied increasingly to better understand the underlying principles of
deep learning algorithms. However, neural representations of distributional
regression models, such as the Cox model, have received little attention so far.
We close this gap by proposing a framework for \longname{} (\acronym), which
includes neural representations of the aforementioned models. We empirically
demonstrate that the neural representations of models in \acronym{} can serve as
a substitute for their classical statistical counterparts in several
applications involving continuous, ordered, time-series, and survival outcomes.
We confirm that models in \acronym{} empirically match the performance of
several statistical methods in terms of estimation of partial effects,
prediction, and aleatoric uncertainty quantification. \acronym{} covers both
interpretable statistical models and flexible neural networks opening up new
avenues in both statistical modeling and deep learning.
\end{abstract}

\section{INTRODUCTION}

Many fundamental statistical modeling approaches, such as random forests or
generalized additive models, focus on predicting the (conditional) mean
\citep{wedderburn_quasi-likelihood_1974,wood2017generalized,breiman2001random}.
While these approaches comes with extensive theoretical guarantees, they largely
ignore aleatoric uncertainty, i.e.,~the stochasticity in the conditional outcome
distribution. Recent developments therefore increasingly model the entire
conditional distribution instead of its conditional mean \citep{KNEIB202399}.
Motivated by their universal approximation property, neural networks based on
Gaussian mixtures were proposed early to learn conditional outcome distributions
\citep{bishop1994mixture}. Approaches to model (all) distributional parameters
of a parametric distribution as a function of features have been proposed in
statistics and have gained more popularity only in recent years \citep[see][for
details]{KNEIB202399}. An alternative to the aforementioned parametric
distributional regression methods was developed in the form of semi-parametric
transformation models, which remove the restrictive assumption of a parametric
outcome distribution by employing a feature-dependent transformation of the
outcome to a simple base distribution
\citep{cheng1995analysis,mclain2013efficient,hothorn2014conditional}. This
modeling approach is closely related to the concept of normalizing flows in deep
learning \citep{rezende2015variational,papamakarios2021normalizing}. 

The idea behind both normalizing flows and transformation models is to learn a
feature-dependent transformation between the outcome and a latent variable with
a fixed, simple distribution, such as the multivariate standard normal
distribution. However, normalizing flows are more expressive than transformation
models due to their non-parametric nature and by relying on deep neural
networks. Transforming an outcome to a well-behaved latent scale goes back to
\citet{box1964analysis} and several works have already pointed out the
connection between normalizing flows and transformation models
\citep{baumann2021deep,sick2021deep,ausset2021surv,kook2022deep}. However, no
formal connection has been established so far.

We close this gap by proposing a class of conditional flows that can be used as
a neural network based substitute for various distributional regression
approaches (transformation, survival, and mixture models) in statistics. 

\paragraph{Our Contribution}   
In this work, we propose an assumption-lean modeling framework termed
\longname{} (\acronym). The proposed framework is built around conditional
flows, or equivalently, a neural and non-parametric variant of a transformation
model replacing its parametric transformation function (i.e., the inverse
conditional flow) with a monotone neural network. Using this model class, we
show how numerous statistical models can be understood as a \acronym. To obtain
interpretable model terms, a neural basis function approach is used for
processing features. Models in \acronym{} can be treated in a unified maximum
likelihood framework, covering continuous, discrete binary, ordered as well as
censored outcomes. We compare models in \acronym{} with state-of-the-art
distributional regression models in real-world applications featuring ordinal,
(clustered) time-series and survival outcomes and demonstrate that \acronym{} is
a competitive alternative. We conclude with a benchmark study in which
\acronym{} is shown to be a well-working, neural network-based framework for
distributional regression in terms of predictive performance. \acronym{} thus
offers one way to interpolate between statistical models and complex neural
networks in terms of flexibility and intelligibility.  

\section{RELATED LITERATURE}

\paragraph{Structured Neural Regression Models}

Structured neural regression bridges the gap between the inherent
interpretability of statistical models and the predictive power of black-box
neural networks as universal approximators \citep{hornik1989multilayer}. Early
attempts to integrate statistical models and neural networks focused on data
exploration \citep{ciampi1995designing}, combining pre-trained statistical
models within a network of models, or using the parameters of statistical models
as initial network weights \citep{ciampi1997statistical}, with extensions to
generalized additive neural networks
\citep{potts1999generalized,de2007generalized}.

Advancements in deep learning, including automatic differentiation and the
availability of efficient modular software libraries, have led to the recent
introduction of semi-structured distributional regression
\citep{rugamer2023semi}. This approach proposes an end-to-end differentiable
hybrid network architecture that combines interpretable structured additive
predictors, as seen in GAMs, and arbitrary deep learning models within a
distributional regression framework. This has led to a series of subsequent
developments, particularly critical in domains such as medicine, where it is
essential to model interpretable effects of tabular features alongside
unstructured data modalities \citep{rudin2018stop}, such as images
\citep{baumann2021deep,dorigatti23a,kook2022deep,kopper2022deeppamm,herzog2023stroke}.

A distinct approach is proposed in neural additive models
\citep{agarwal2021neural} and its extensions
\citep[e.g.,][]{changnode,radenovic2022neural,yang2021gami}. Here, a standard
additive GAM predictor for the conditional mean is assumed, with the shape
functions for each feature learned in separate subnetworks. This approach
overcomes the potential limitations of pre-defined basis functions to model
highly complex or jagged functions, albeit at the cost of drastically increased
parameter counts and a still-evolving theoretical foundation
\citep{heiss2019implicit,zhang2022deep}. Despite the recent progress in
structured neural regression models, they typically impose the restrictive
assumption of a known parametric outcome distribution, highlighting the
importance of non-parametric normalizing flows. 

\paragraph{Normalizing Flows}

Normalizing flows model a random variable $Y$ with a complex distribution
through compositions of invertible and differentiable transformations of a
latent random variable $\varepsilon$ that has a known base distribution with no
free parameters, such as a standard normal distribution. In our context, these
diffeomorphisms are parameterized by deep neural
networks~\citep{dinh2016density}. Specifically, normalizing flows express $Y$ as
$\phi(\varepsilon)$, where $\phi : \mathbb{R} \to \mathcal{Y}$ is a flow.
Usually, $\phi$ follows a pre-specified functional form that enables fast
inversion and computation of the determinant of the Jacobian, which is required
to obtain the density of $Y$. Examples include coupling~\citep{dinh2016density},
planar and radial~\citep{rezende2015variational},
autoregressive~\citep{kingma2016improved}, or residual
flows~\citep{chen2019residual}. The weights parametrizing $\phi$ can be learned
through maximum likelihood training, and several such transformations can be
stacked to approximate arbitrarily complex distributions. Normalizing flows have
been successfully applied both to conditional and unconditional generative
modeling, and distributional
regression~\citep{papamakarios2021normalizing,winkler2019learning}.

\section{ASSUMPTIONS IN STATISTICAL MODELING} \label{sec:ibs}

Statistical modeling requires direct input by the data analyst in the form of
assumptions that express relevant domain knowledge, which is unavoidable in
situations with scarce or highly complex data such as multi-task
learning~\citep{silver2013lifelong}, biology~\citep{xu2019machine}, material
science~\citep{childs2019embedding}, physics~\citep{stewart2017label}, and more.
Common assumptions fall into two distinct categories.

\paragraph{Distributional Assumptions} The construction of both mean and
distributional regression models usually requires assuming a known, parametric
family of distributions of the underlying outcome. Typical examples in
statistical modeling encompass the linear model (Gaussian error distribution),
generalized linear and additive models assuming an exponential family
\citep{hastie1986generalized,nelder1972}, or structured additive distributional
regression models, specifying additive predictors for all distribution
parameters of an \textit{a priori} known parametric distribution
\citep{KNEIB202399}. Identifying an appropriate parametric distribution for the
model generally depends on either solid domain knowledge or exhaustive model
comparisons. Moreover, most distributional assumptions do not allow for
multimodality in the learned distribution. Unmet assumptions can lead to
inconsistent, biased, or inefficient estimation of model parameters \citep[see,
e.g.,][]{pawitan2001all,White1982}. To robustify results against misspecified
distributions, various approaches try to compensate for unexplained variance or
samples, e.g., by applying outlier removal or using a more heavy-tailed
distribution \citep[see, e.g.,][]{huber2011robust}. While this practice can
improve performance, model validation and diagnostics are manual and iterative
processes that, in turn, often require domain knowledge
\citep{white1981consequences}. 

\paragraph{Structural Assumptions} To foster interpretability and limit
complexity, statistical models commonly make additional structural assumptions.
Two of the most common ones are additivity and linearity of predictors. This
means that the conditional mean of a response $Y$ given features $X$, $\mu(X)
\coloneqq \mathbb{E}[Y| X]$ (or any other aspect of the conditional
distribution), relates to features by $g(\mu(X)) = X\beta$ with invertible link
function $g$ and feature weights $\beta$. One of the most prominent examples
following this assumption is the generalized linear model \citep{nelder1972}.
Extensions of (generalized) linear models, such as GAMs
\citep{hastie1986generalized,wood2017generalized}, allow to go beyond linear
feature effects by using, e.g., a spline basis representation to introduce
non-linearity. In this case, domain knowledge is often needed to choose the
best-fitting spline basis, the number and position of knots, or the amount of
smoothness \citep[see,
e.g.,][]{gu2013smoothing,schumaker2007spline,wood2017generalized}. Another
typically human-based decision for such models is the inclusion of higher-order
feature interactions. Limiting the degree and number of interactions allows
controlling the number of parameters (and thus scalability) while ensuring a
certain level of interpretability. 

\section{INVERSE CONDITIONAL FLOWS FOR DISTRIBUTIONAL REGRESSION}

Consider observations $\{(y_i, x_i)\}_{i=1}^n$ of a univariate outcome $Y \in
\calY \subseteq \RR$ and features $X \in \calX$. In this work, we focus on
modeling the entire conditional distribution of $Y$ given $X$. We propose a
flexible class of models for the conditional distribution of $Y$ given $X$ that
interpolates between highly flexible normalizing flows (low domain knowledge)
and parametric models (high domain knowledge) for various outcome types. In this
class, a model needs two components to be fully specified. First, a
parameter-free base distribution with cumulative distribution function (CDF) $F
: \RR \to [0, 1]$ and continuous, two times differentiable, log-concave density
$f$; and second, a conditional flow $\phi : \RR \times \calX \to \calY$ which
maps observations from the parameter-free base distribution to the conditional
outcome distribution for all constellations of features. By conditional flow, we
refer to a (composition of) function(s), later parameterized by neural networks,
which is monotonically increasing for all possible realizations of the features.

Let $\mathcal{P}$ denote the class of all conditional CDFs with sample space
$\calY$ and conditional on features in $\calX$. Then for each base CDF $F$, the
class of models under investigation can be defined as the set $\Phi_F$
containing all conditional flows $\phi: \RR \times \calX \to \calY$ such that
for all conditional distributions $F_{Y|X} \in \mathcal{P}$, we have that for
$\epsilon \sim F$, $\phi(\epsilon, X) \sim F_{Y |X}$.

Domain knowledge can now enter as restrictions on $\Phi_F$. The conditional
cumulative distribution function of $Y$ given $X$, denoted by $F_{Y|X}$, can for
all $x \in \calX$ be written as 
\begin{equation*}
    F_{Y | X = x}(\bcd) = F(\phi^{-}(\bcd, x)),
\end{equation*}
where $\phi^{-}(\bcd, x) \coloneqq \sup \{z \in \RR : \phi(z, x) \leq \bcd\}$
denotes the conditional (generalized) \emph{inverse} flow. The inverse flow
plays an important role in training models in \acronym{} (see
Section~\ref{sec:training}). Next, we consider what types of assumptions can be
imposed on these models and how those assumptions affect model capacity,
i.e.,~the flexibility of conditional flows contained in $\Phi_{F}$ and the
conditional distributions they can model (see Figure~\ref{fig:overview} for an
example). Then, we discuss explicit parameterizations of and how to train models
in \acronym{}.

\subsection{Assumptions on the Base Distribution}

On their own, assumptions on the base distribution do not limit model capacity,
because any conditional CDF $F_{Y | X} \in \mathcal{P}$ can be composed as $F
\circ F^{-1} \circ F_{Y | X}$. Then the set of functions 
\begin{equation*}
\{\phi^- = F^{-1} \circ F_{Y | X} \mid F_{Y|X} \in \mathcal{P}\}
\end{equation*}
gives rise to all conditional flows with base CDF $F$.
Choosing a particular $F$ only fixes the scale on which to interpret the
components of the flow $\phi$.

\begin{example}[Assumptions on the base distribution]
Binary classification via logistic regression can be thought of as \acronym{}
with standard logistic CDF $F$ and inverse conditional flows on the log-odds
scale, i.e., 
\begin{equation*}
\phi^-(y, x) = \log\frac{F_{Y|X = x}(y)}{1 - F_{Y | X =x}(y)}.
\end{equation*}
However, the conditional cumulative distribution $F_{Y|X}$ can be modeled with
other base distributions, such as the standard minimum extreme value, or
standard normal distribution. These correspond to inverse conditional flows
interpretable on the log-hazard $\operatorname{cloglog}(\pi) =
\log(-\log(1-\pi))$, or probit $F_{N(0,1)}^{-1}$ scale
\citep{tutz2011regression}.
\end{example}

\subsection{Structural Assumptions}

\begin{figure}[!t]
\centering
\includegraphics[width=0.75\columnwidth]{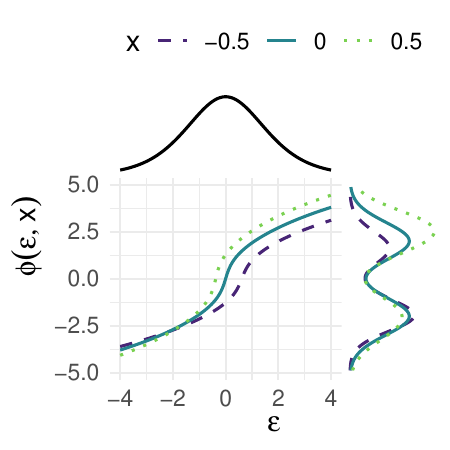}
\caption{
Depiction of the location-scale \acronym{} in Example~\ref{exmpl:drift}. For
three values of $X$ (dashed/solid/dotted), the standard logistic base
distribution (top side) is transformed into the conditional outcome distribution
(right side) via the conditional flows (middle). In this example, the
distribution for $x=0$ is a normal mixture with equal weights (solid line). 
}\label{fig:overview}
\end{figure}

Structural assumptions take a variety of forms and limit model capacity more or
less severely. In \acronym{}, we impose an additivity assumption on the
conditional flow $\phi$ in terms of the features, which is the distribution-free
analog to additivity assumptions in GAMs on the scale of the conditional mean.
We define the class of location-scale conditional flows used in \acronym{} by
\begin{equation} \label{eq:addflow}
    \Phi^{\text{L-S}}_{F} = \{\phi \in \Phi_{F} \mid \ \phi(\varepsilon, x) =
    \phi_0(\mu(x) + \sigma(x)\varepsilon)\},
\end{equation}
where $\mu : \calX \to \RR$ controls location, $\sigma : \calX \to \RR_+$
controls scale, and $\phi_0 : \RR \to \calY$ is called \emph{reference flow},
because it is the flow from $F$ to any $F_{Y | X = x_0}$ for which $\mu(x_0) =
0$ and $\sigma(x_0) = 1$ (see Figure~\ref{fig:overview}). Additivity assumptions
on $\phi$ restrict model capacity but do not imply a fixed family of conditional
outcome distributions because no distribution assumptions are implied by the
reference flow. Features can only change location and scale of the base
distribution before applying the reference flow
\citep{rezende2015variational,siegfried2022distribution}. 
\begin{example}[Structural assumptions]\label{exmpl:drift}
Here, we illustrate how to construct and sample from a location-scale
conditional flow with a single feature. First, we choose a reference flow
$\phi_0 \coloneqq F_{Y | X = x_0}^{-1} \circ F$ for an arbitrary reference $x_0
\in \calX$. Then, we introduce a shift $\mu$ and scale effect $\sigma$. Samples
from $Y | X = x$ are then generated via $Y := \phi_0(\sigma(x)\epsilon +
\mu(x))$. For example, we choose $\epsilon \sim F$ to follow a standard logistic
distribution, and a Gaussian mixture $0.5 N(-2, 1) + 0.5 N(2, 1)$ for $F_{Y | X
= x_0}$. Further, we introduce a nonlinear $\mu : x \mapsto \mu(x)$ with $\mu(x)
= \exp(1 - \exp(-x)) - 1$ and $\sigma : x \mapsto \sqrt{\exp(x)}$ as the shift
and scale effects. Here, $\mu$ and $\sigma$ are such that the reference is $x_0
= 0$ since $\mu(x_0) = 0$ and $\sigma(x_0) = 1$. The interplay between base
distribution, conditional flow and conditional outcome distribution in
\acronym{} is shown in Figure~\ref{fig:overview}.
\end{example}
A combination of distributional and stronger structural assumptions can fix the
conditional outcome distribution and thus severely limit model capacity. When
limiting the reference flow to the identity, i.e.,~
\begin{equation*}
    \{\phi \in \Phi_{F}^{\text{L-S}} \mid \phi(\varepsilon, x) = \mu(x) +
    \sigma(x) \varepsilon\},
\end{equation*}
the distribution of $Y | X$ is limited to the location-scale family induced by
$F$. For instance, with $\epsilon \sim N(0, 1)$, $\calY \subseteq \RR$ and
$\calX \subseteq \RR^d$, $\mu : x \mapsto (1, x)^\top\beta$ and $\sigma \equiv c
> 0$, such linear conditional flows recover linear regression.

\subsection{Training Models in \acronym{} via Maximum Likelihood}\label{sec:training}

Models in the \acronym{} framework lend themselves to estimation via maximum
likelihood. \acronym{} can be used to model the distribution of outcomes with
binary, ordinal, count-valued, continuous and mixed discrete-continuous sample
space $\mathcal{Y}$. For continuous (exact) responses, the likelihood function
is given by the log-density, whereas for discrete responses, the likelihood is
obtained as a difference in cumulative distribution functions. Since we have
access to the entire conditional distribution, we can also evaluate the
likelihood contributions of censored outcomes.

We consider the log-likelihood function for exact continuous, discrete and
uninformatively censored outcomes. For exact continuous observations $y \in
\RR$, the log-likelihood $\ell : \Phi_F \times \calY \times \calX \to \RR$ is
given by the log-density, i.e.,~
\begin{equation*}
\ell(\phi, y, x) = \log f(\phi^-(y, x)) \frac{\dd}{\dd \upsilon}\phi^-(\upsilon,
x) \big\rvert_{\upsilon = y}.
\end{equation*}
Using deep learning libraries, involved gradients of the log-likelihood can be
computed efficiently. For discrete outcomes supported on $\{y_1, y_2, \dots,
y_K\} \subseteq \RR$, we have for $k \in \{1, \dots, K\}$, 
\begin{equation*}
    \ell(\phi, y_k, x) = \log [F(\phi^-(y_k, x)) - F(\phi^-(y_{k-1}, x))],
\end{equation*}
where, by convention, $F(\phi^-(y_0, x)) = 0$ and $F(\phi^-(y_K, x)) = 1$.
The log-likelihood of interval censored outcomes $(y_l, y_u]$ can be defined
likewise, 
\begin{equation*}
\ell(\phi, (y_l, y_u], x) = \log [F(\phi^-(y_u, x)) - F(\phi^-(y_l, x))],
\end{equation*}
with a slight abuse notation when allowing interval-valued observations.

To evaluate the likelihood, we thus need evaluate the base CDF $F$ and the
(generalized) inverse flow $\phi^-$. Since $F$ is fixed, we now turn to
parameterizations of $\phi^-$. 

\subsection{Parameterizing Models in \acronym{}}
%
%
We parameterize models in \acronym{} explicitly via neural networks. Three
components need to be specified: The inverse reference flow $\phi_0^-$, shift
$\mu$, and scale $\sigma$ effect. The inverse reference flow needs to be
monotonically increasing and its smoothness depends on the outcome type.
Location effects are unconstrained, whereas scale effects need to fulfill a
simple positivity constraint. In this work, we parameterize all three functions
via neural networks.

\paragraph{Inverse Reference Flow} For discrete outcome types, a dummy encoded
basis with increasing coefficients is sufficient to ensure monotonicity. For
absolute continuous outcomes, $\phi_0^-$ can be a smooth invertible function.
Classically, $\phi_0^-$ has been parameterized via basis expansions, such as
$B$-splines \citep{hothorn2014conditional} or polynomials in Bernstein form
\citep{hothorn2018most,mclain2013efficient}. Here, we parameterize the reference
flow via monotonic neural networks \citep{huang2018neural}. Sufficient
conditions for monotonicity are given in the following result.

\begin{proposition}[Monotonicity of the conditional flow] \label{prop:mnn}
Consider an inverse conditional flow of the form 
\begin{equation*}
\phi^{-}(y, \mathbf{x}) = \phi^{-}_{y\mathbf{x}}(\phi^{-}_{y}(y),
\varphi(\mathbf{x})),
\end{equation*}
where $\phi^{-}_{y\mathbf{x}}$, $\phi^{-}_{y}$, and $\varphi$ are feed-forward
neural networks. For $\phi^{-}(y, \mathbf{x})$ to be strictly monotonically
increasing in $y$, it is sufficient for $\phi^{-}_{y\mathbf{x}}$ and
$\phi^{-}_{y}$ to have strictly positive weights and strictly monotonic
activation functions (e.g., tanh activations).
\end{proposition}

In the special case of parameterization \eqref{eq:addflow}, we thus only need to
choose $\phi_0^-$ to be a monotonic neural network. The proof of the more
general result can be found in, e.g., \cite{Silva2018TowardsCE}.

\paragraph{Location and Scale Effects}

To avoid restrictive structural assumptions while preserving interpretability,
we specify predictors $\psi$ for $\mu$ and $\sigma$ using neural basis functions
\citep{agarwal2021neural}, i.e.,
\begin{equation} \label{eq:pred}
    \psi(x) = \textstyle\sum_{j=1}^J \rho_j(x_j),
\end{equation}
where each $\rho_j$ represents a feature-specific network learning an adaptive
basis function for the respective feature $x_j$. This network can be further
extended to, e.g., include bivariate feature effects $\sum_{i,j: i\neq j}
\rho_{i,j}(x_i,x_j)$ as also done in our numerical experiments or even
higher-order interactions. In case multiple feature effects contain the same
feature $x_j$, various approaches exist to ensure the model's identifiability
\citep[see, e.g.,][]{rugamer2023semi}. Identifiability is particularly important
if the model predictor in \eqref{eq:pred} is further extended by a more complex
(deep) neural network capturing higher-order interaction effects. In that case,
the recently proposed approach in \citet{pho2023} provides a non-invasive
post-hoc adaption of the model that is also suitable for our approach.

\section{NUMERICAL EXPERIMENTS}

We now present a variety of numerical experiments where we investigate whether \acronym{}
is a viable substitute to one or more established statistics approaches of
similar complexity and aligns with their goodness-of-fit, effect estimation, and
predictive performance. These experiments also provide insights into whether
normalizing flows can be similarly interpretable as statistical models. In the
Supplementary Material, we further analyze the hyperparameter stability of
models in \acronym{} and give the explicit parameterization of all models and
competitors used in the experiments.

\paragraph{Setup} In Sections~\ref{sec:wine}--\ref{sec:elec}, and~\ref{sec:dr},
we parameterize $\phi_0^-$ in terms of an invertible neural network. To further
demonstrate the ease with which to interpolate between a fully neural and
semi-parametric $\phi_0^-$, we specify $\phi_0^-$ in
Sections~\ref{sec:multimodal} and~\ref{sec:surv} using polynomials in Bernstein
form, an alternative to monotone neural networks used in transformation models
\citep{hothorn2018most} and also recently discussed for normalizing flows
\citep{Ramasinghe2022}.

\subsection{Ordinal Regression} \label{sec:wine}
The UCI ``Wine quality'' dataset \citep{cortez2009modeling} contains 1599 red
wines whose quality is described on an ordinal scale (10 levels of which only
3--8 have been observed). We consider five features, namely fixed and volatile
acidity, citric acid and residual sugar content, and concentrations of
chlorides. Non-linear effects are specified by a feature-specific ReLU-network
with four hidden layers and eight units each. 
\begin{figure}[!ht]
    \centering
    \includegraphics[width=\columnwidth]{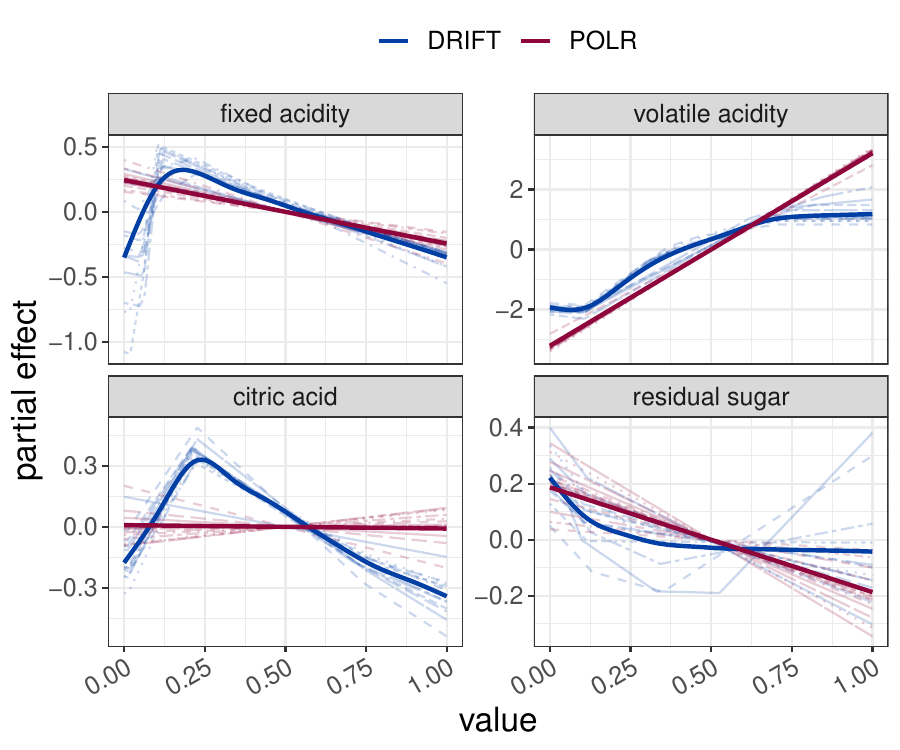}
    \caption{%
    Estimated partial effects for four features in a 20-fold cross-validation of
    the UCI wine quality dataset using a \acronym{} and a proportional odds
    logistic regression (POLR) model.
    }%
    \label{fig:wine}
\end{figure}
\begin{figure*}[!ht]
   \centering
   \includegraphics[width=0.47\textwidth]{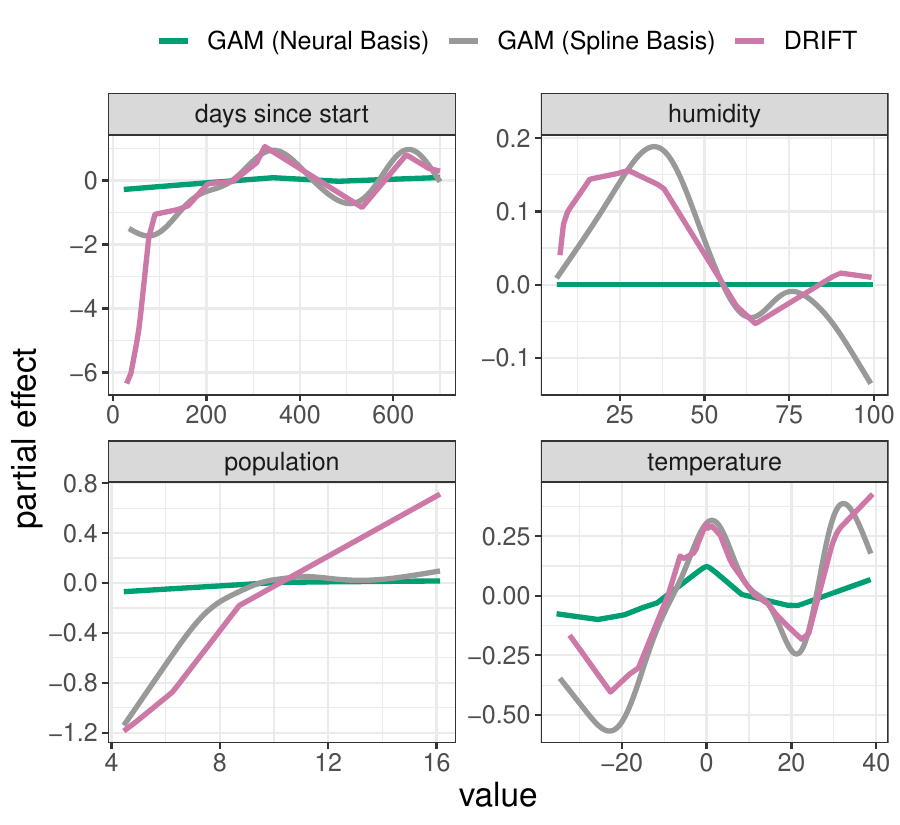} 
   \hfill
   \includegraphics[width=0.47\textwidth]{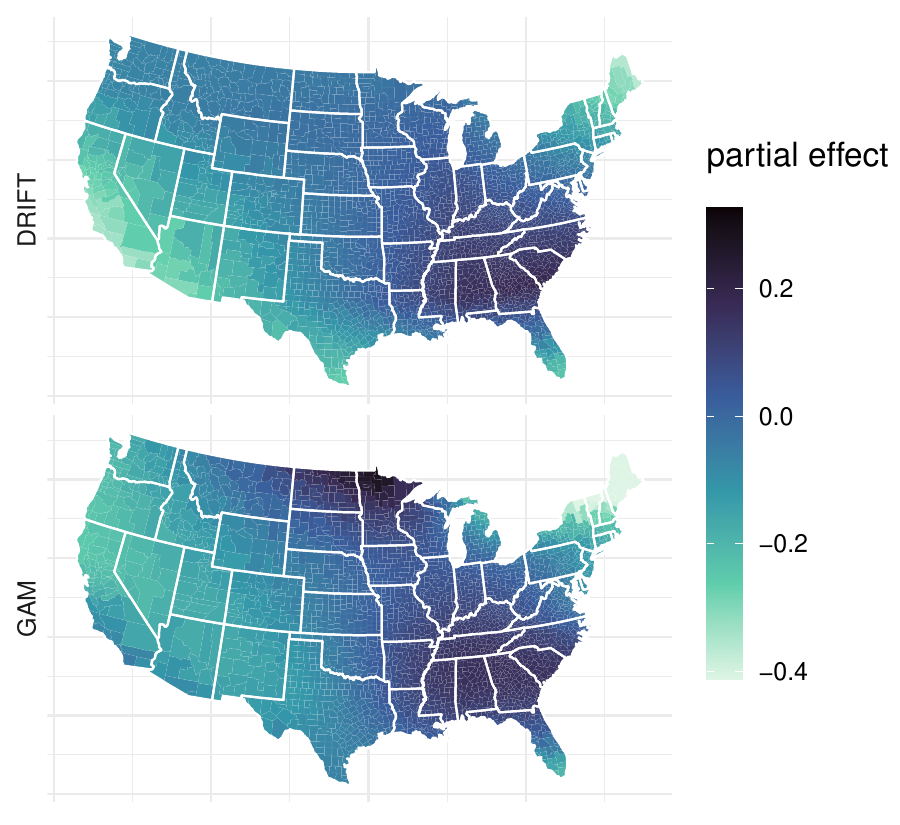} 
    \caption{%
    Left: Estimated effects on the prevalence of Covid-19 using a GAM with
    neural or spline basis and \acronym{} (colors). Right: Estimated spatial
    effects on the prevalence of Covid-19 from \acronym{} and GAM (with spline
    basis).
    }\label{fig:covid}
\end{figure*}

\textbf{Results}: In a 20-fold cross-validation, an ordinal \acronym{} performs
on par with the standard proportional odds logistic regression with linear
feature effects (log-score $-$1.11 (0.06) \emph{vs.} $-$1.12 (0.06)) and partial
effect estimates deviate from the proportional odds model (POLR,
Figure~\ref{fig:wine}).

\subsection{Generalized Additive Models} \label{sec:gamnam}

To compare \acronym{} with GAMs, we re-analyze a spatio-temporal data set of
number of Covid-19 incidences in the US previously analyzed in \cite{pho2023}.
We follow their data cleaning procedure and model the prevalence of infections
using a GAM (either by using a B-spline basis or feature-specific neural
networks) with features for population, date, latitude and longitude,
temperature, and humidity. We then compare these previous methods with
\acronym{} (using a monotone neural network for $\phi_0^-$) qualitatively by
analyzing the estimated partial effects. 

\textbf{Results}: Inspecting Figure~\ref{fig:covid} (left) we find that the
partial effects based on neural basis functions are underfitted compared to
spline-based partial effects, and even collapse to a zero effect for humidity.
This difficulty in training neural additive models is a well-known phenomenon in
the literature. In contrast, effects estimated via \acronym~look very similar to
those obtained from a traditional GAM with spline basis. This is also the case
for the spatial effect (Figure~\ref{fig:covid}, right). 

\subsection{Time Series Analysis} \label{sec:elec}

We forecast the hourly electricity consumption of 370 clients contained in the
UCI \textit{Electricity} dataset \citep{Dua.2019}. We model each of the
univariate time series with 48 consecutive lags based on 9 days of data
(starting 2014-07-01). We first determine the optimal number of training epochs
using data for the first 7 days as training set and the 8-th day as validation
set for early stopping. Then, knowing the optimal number of epochs, we use the
first 8 days for the final training and forecast on the 9-th day. 

\textbf{Results}: We obtain a log-score of -0.538 (0.195). As a comparison, we
use the \textit{auto.arima} function \citep{Hyndman.2023} that fits an ARIMA
model with an automatic search for the best model parameters. This results in a
worse performance with a log-score of -4.434.

\subsection{Mixture Modeling for Multimodal Distributions}
\label{sec:multimodal}

We next demonstrate the flexibility of our approach to model multimodal
distributions. To this end, we investigate the ATM dataset from
\cite{Ruegamer.2023}, known to follow a time-dependent process with
mode-switching behavior. We focus on the multimodality in the data and compare
our approach against mixture models -- a classical choice for multimodal data.
Both approaches use the time information as a feature to allow for a
time-varying density estimation. 
\begin{figure}[!ht]
    \centering
    \includegraphics[width=\columnwidth]{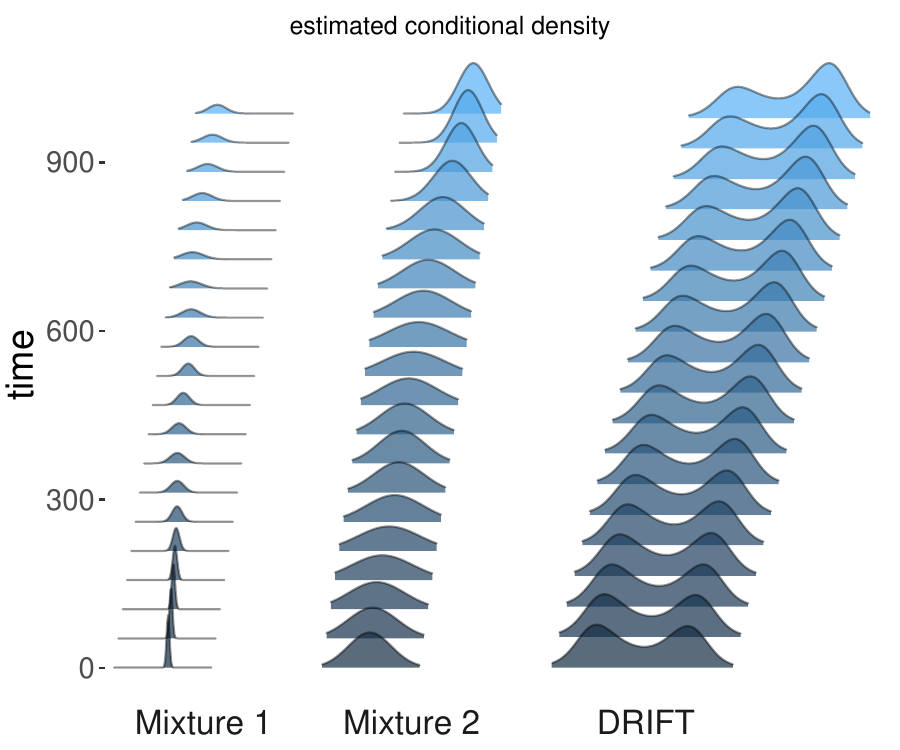}
    \caption{%
    Left and center: Estimated Gaussian densities over time (y-axis) for the two
    mixture components in the mixture model. Right: Estimated densities over
    time using \acronym.
    }\label{fig:yeast}
\end{figure}

\begin{figure*}[t!]
    \centering
    \includegraphics[width=\textwidth]{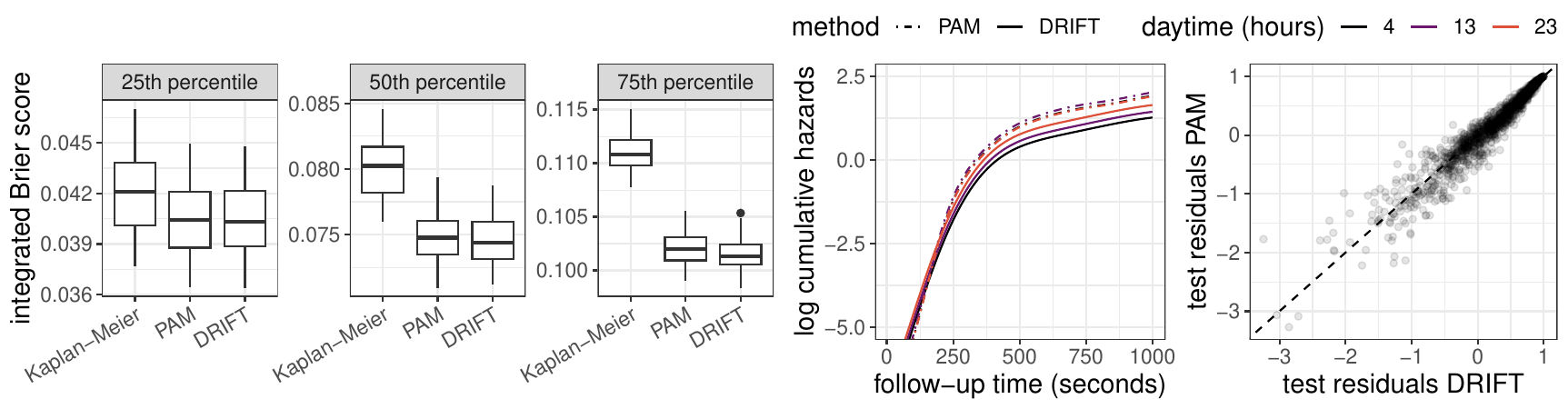}
    \caption{%
    Left: Predictive performance in terms of integrated Brier score (lower is
    better, evaluated at the 25th, 50th, and 75th percentile) of a Kaplan-Meier
    estimator, a piece-wise exponential additive model (PAM) and a \acronym{}.
    Middle: Estimated log cumulative hazards given daytime (in hours; colors).
    Right: Out-of-sample martingale residuals show comparable prediction errors
    for DRIFT and PAM.
    }\label{fig:surv}
\end{figure*}

\textbf{Results}: Based on a predefined test dataset, we find that the negative
log-likelihood (smaller is better) of our approach is 2.09 whereas the mixture
model with 2 components as reported in results in a value of 2.27. When
investigating the learned densities (see Figure~\ref{fig:yeast}), we find that
both approaches capture the 2 modes at later time points, but \acronym{} works
slightly better by also capturing the multimodality at earlier points in time.
Furthermore, our approach does not require to specify the number of modes
\textit{a priori}.

\subsection{Survival Analysis} \label{sec:surv}

Next, we use a \acronym{} to learn spatio-temporal determinants of response
times (time-to-arrival) of the London fire brigade to fire-related emergency
calls. The data has been previously used in
\citet{taylor2017spatial,kopper2022deeppamm}. \acronym{} allows various survival
analysis model classes to be used, e.g.,~a piecewise exponential additive model
\citep{bender2018} or a Cox proportional hazards \citep{cox1972regression}
model. Using 25 different hold-out splits we compare a \acronym{} resembling a
Cox model with smooth log cumulative baseline hazard $\phi_0^-$ (see
Supplementary Material for details) with a piecewise exponential additive model
based on the reweighted integrated Brier score
\citep[IBS,][]{sonabend2022scoring}. We also review the performance of a
featureless learner (Kaplan-Meier) and report the IBS at the quartiles of the
follow-up time.

\textbf{Results}: Figure~\ref{fig:surv} shows the comparison between results
obtained by a piecewise exponential additive model (PAM) and \acronym{} in terms
of prediction performance (integrated Brier score; left) and qualitatively in
terms of estimated log cumulative hazard functions conditional on time of day.
The \acronym{} shows slightly better out-of-sample performance in terms of
re-weighted integrated Brier score and estimtates a stronger influence of time
of day on the survivor curve compared to the piecewise exponential additive
model. Martingale residuals \citep{barlow1988residuals} on the held-out data for
a single split (Figure~\ref{fig:surv}, right) show that PAM and the \acronym{}
make qualitatively similar prediction errors and illustrate that \acronym{}
allows residual-based model checks for non-continuous outcomes.

\subsection{Structured Additive Distributional Regression and Transformation
Models: Benchmark study} \label{sec:dr}

Finally, we check if \acronym{} is able to match the predictive performance of
other additive distributional regression approaches. For the comparison, we use
a structured additive distributional regression with parametric distribution
assumption \citep{klein_bayesian_2015}, a transformation model
\citep{hothorn2014conditional}, and \acronym{} for which the reference flow is
non-parametric. As both structured additive distributional regression and
transformation models represent very flexible approaches that are closely
related to our method, we run a benchmark study to compare performances using an
extended collection of the classical UCI machine learning repository datasets
\citep{Dua.2019}. As base distribution $F$ we use a Gaussian for both \acronym{}
and the transformation model. The distributional regression is also defined
based on a Gaussian distribution. The predictors for the transformation and
distributional regression model are defined using thin-plate regression splines.
For \acronym{} we use neural basis function splines based on one fixed
architecture (see Supplementary Material for details). For $\phi^-_0$, we use a
simple monotonic neural network with two hidden layers of either 10 or 20
neurons each. We also compare these methods when using a semi-structured
predictor, i.e., when enhancing the structured predictor with a deep neural
network for all features and methods (see details in
Appendix~\ref{app:benchmark}).
\begin{table}[!ht]
\centering
\caption{\footnotesize Comparison results for different datasets (rows) and the
\emph{structured} methods (DR: distributional regression; TM: transformation
model, \acronym{}: Location-scale) showing the mean log-score (and standard
deviation in brackets) based on a 10-fold cross-validation. The best methods per
dataset are highlighted in bold.
} \label{tab:benchmark}
\small
\begin{tabular}{lrrr}
\toprule
Dataset & DR & TM & DRIFT \\
\midrule
Airfoil & $-$3.6 (0.3) & $-$3.2 (0.2) & \textbf{$-$3.1} (0.1) \\
Concrete & $-$3.4 (0.2) & $-$3.5 (0.3) & \textbf{$-$3.3} (0.1) \\
Diabetes & $-$5.8 (0.3) & $-$5.6 (0.4) & \textbf{$-$5.2} (0.2) \\
Energy & $-$2.7 (0.2) & $-$2.6 (0.1) & \textbf{$-$2.3} (0.1) \\
Fish & $-$1.4 (0.2) & \!\textbf{$-$1.3} (0.3) & \textbf{$-$1.3} (0.1) \\
ForestF & $-$1.9 (0.3) & $-$1.7 (0.2) & \textbf{$-$1.4} (0.2) \\
Ltfsid & $-$6.5 (0.2) & $-$6.2 (0.3) & \textbf{$-$4.7} (0.1) \\
Naval & \phantom{$-$}4.0 (0.2) & \phantom{$-$}3.8 (0.3) & \phantom{$-$}\textbf{5.1} (0.1) \\
Real & $-$1.3 (0.4) & $-$1.1 (0.2) & \textbf{$-$0.9} (0.1) \\
Wine & \phantom{$-$}0.5 (0.2) & \phantom{$-$}0.8 (0.4) & \phantom{$-$}\textbf{4.2} (0.2) \\
Yacht & $-$1.5 (0.3) & $-$1.7 (0.4) & \textbf{$-$0.8} (0.1) \\
\bottomrule
\end{tabular}
\end{table}

\textbf{Results}: Our results (Table~\ref{tab:benchmark}) confirm that
\acronym{} is able to match the performance of other methods, in many cases even
outperforming them. This is particularly notable for datasets in which the
outcome exhibits a non-Gaussian distribution (unconditionally). For example, in
the \emph{Wine} dataset, the outcome is technically discrete, but commonly
treated as continuous. Here, distributional regression with a parametric
Gaussian assumption yields the worst results. Using a transformation model can
improve this result, however, both parametric alternatives are outperformed by
the non-parametric neural reference flow used in \acronym{}.

We also run the same comparison as presented in Table~\ref{tab:benchmark} when
using model formulations discussed in \citet{baumann2021deep, pho2023}, namely
(i) when combining structured predictors with neural networks
(Table~\ref{tab:benchmark3}), (ii) neural basis functions for all three
approaches (Table~\ref{tab:benchmark2} in the Supplement), and (iii) only deep
neural network predictors for all three methods (Table~\ref{tab:benchmark1} in
the Supplement). Similar to the results presented in Table~\ref{tab:benchmark},
\acronym{} is on par with or improves upon DR and TM.
\begin{table}[t!]
\centering
\caption{\footnotesize Comparison results for different datasets (rows) and
\emph{semi-structured} methods (DR: distributional regression; TM:
transformation model, \acronym{}: Location-scale) showing the mean log-score
(standard deviation) based on a 10-fold cross-validation. The best methods per
dataset are highlighted in bold.}\label{tab:benchmark3}
\small
\begin{tabular}{lrrr}
\toprule
Dataset & DR (Semi) & TM (Semi) & DRIFT (Semi) \\
\midrule
Airfoil & \textbf{-2.9} (0.1) & -3.0 (0.1) & -3.1 (0.5) \\
Concrete & -3.3 (0.1) & -3.3 (0.3) & \textbf{-3.0} (0.3) \\
Diabetes & -5.7 (0.5) & -6.0 (0.4) & \textbf{-5.4} (0.2) \\
Energy & -2.9 (0.1) & -2.7 (0.5) & \textbf{-2.2} (0.1) \\
Fish & \textbf{-1.3} (0.1) & -1.5 (0.2) & \textbf{-1.3} (0.2) \\
ForestF & -2.0 (0.3) & -1.9 (0.2) & \textbf{-1.4} (0.4) \\
Ltfsid & -7.7 (7.5) & -5.9 (0.7) & \textbf{-4.6} (0.1) \\
Naval & \phantom{-}4.1 (1.0) & \phantom{-}3.9 (0.1) & \phantom{-}\textbf{5.1} (0.3) \\
Real & -1.4 (0.4) & -1.4 (0.6) & \textbf{-1.2} (1.1) \\
Wine & -0.2 (0.0) & -0.4 (1.0) & \phantom{-}\textbf{2.1} (2.1) \\
Yacht & -1.0 (0.2) & -2.2 (2.0) & \textbf{-0.5} (0.2) \\
\bottomrule
\end{tabular}

\end{table}

\section{CONCLUSION}

We demonstrate that numerous statistical models can be expressed in the
\acronym{} framework. Equipped with neural basis functions, \acronym{} enables
interpretable model terms with little requirement for manual input from the
modeler. The versatility and practical applicability of \acronym{} is reinforced
by favorable benchmark comparisons and applications involving various outcome
types. Overall, our results suggest that \acronym{} serves as a competitive
neural network-based framework for distributional regression tasks. A promising
avenue for future research involves developing statistical inference methods
tailored to \acronym{}. 

\bibliography{bibliography}


\newpage

\onecolumn

\title{How Inverse Conditional Flows Can Serve as a Substitute for\\
Distributional Regression (Supplementary Material)}

\maketitle

\appendix

\section{ADDITIONAL EXPERIMENTS}

\subsection{Multiple Defaults} \label{app:mult}

\begin{figure}[!ht] 
\centering
\includegraphics[width=\textwidth]{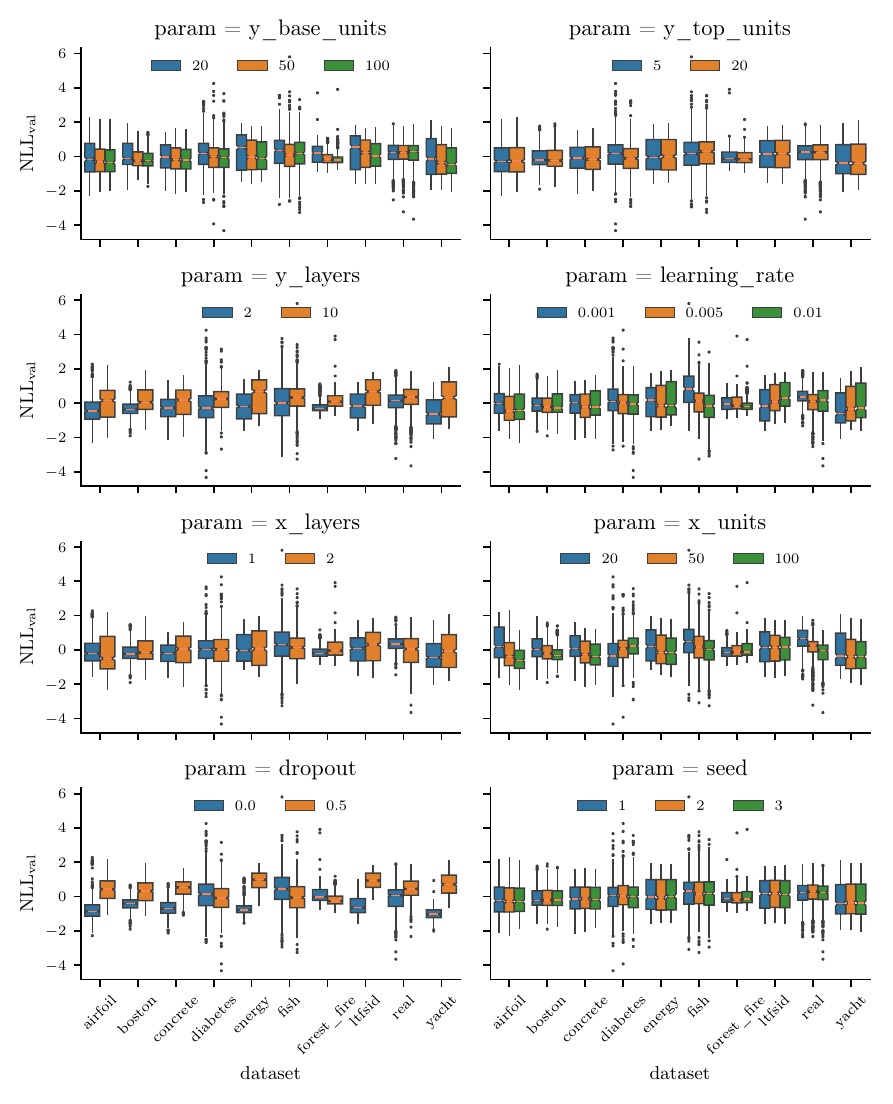}
\caption{The results of the grid search on different datasets.}
\label{fig:gridsearch}
\end{figure}

To investigate the influence of hyperparameters on the performance of
\acronym{s}, a grid search is conducted for a large collection of datasets from
the UCI repository (details below). The purpose of this study is to find a good
default that works well on most datasets such that \acronym{s} can be used
off-the-shelf similar to other (mostly tuning-free) distributional regression
approaches. To account for the stochasticity of the training process, each
combination of hyperparameters is trained 3 times with different seeds. Next to
different learning rates and dropout rates, the architectures of the neural
networks are varied via the number of units and layers for both the features and
outcome of the network (i.e.,~the experiments assume an unstructured predictor
for the feature part of the \acronym{}). The following hyperparameters are
chosen for the grid search: 
\begin{itemize} 
    \item learning rate $\in \{10^{-2}, 5\times 10^{-3}, 10^{-3}\}$,
    \item dropout $\in \{0, 0.5\}$,
    \item seed $\in \{1, 2, 3\}$,
    \item units feature network $\in \{20, 50, 100\}$,
    \item number of layers feature network $\in \{1, 2\}$,
    \item units $\phi^{-}_0 \in \{20, 50, 100\}$,
    \item layers $\phi^{-}_0 \in \{2, 10\}$,
    \item last layer units $\phi^{-}_0\in \{5, 20\}$.
\end{itemize}
The normalized validation negative log-likelihood ($\text{NLL}_{\text{val}}$)
for each choice of hyperparameters across data sets resulting from the grid
search is displayed as boxplots in Figure~\ref{fig:gridsearch}. It is evident
that neither the specific choice of hidden units, the number of layers, nor the
learning rate had a consistent effect on the validation loss, with only an
increased dropout rate leading to slightly worse results across most datasets.
The analysis suggests that the model performance is largely robust with respect
to these hyperparameters. 

\subsection{Influence of Initialization} \label{app:init}
\begin{figure}[!ht] \centering
\includegraphics[width=0.99\textwidth]{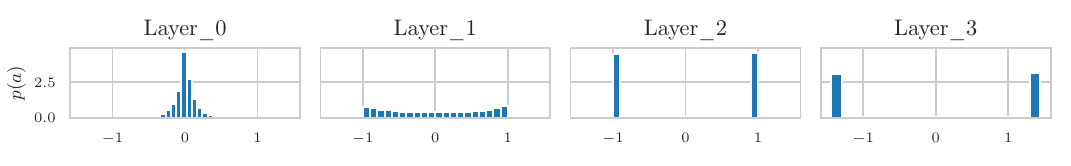}
\includegraphics[width=0.99\textwidth]{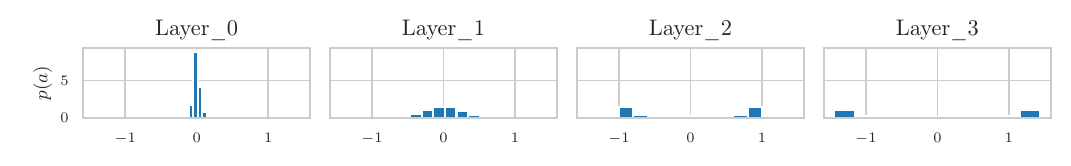}
\includegraphics[width=0.99\textwidth]{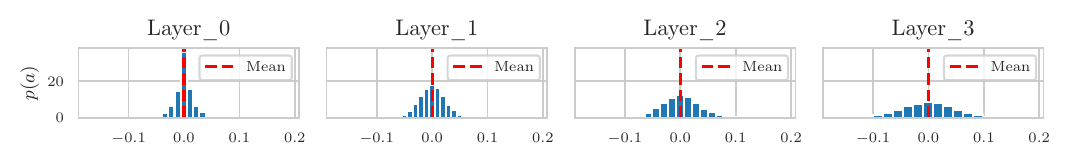}
\caption{%
The value distribution of activations after each layer when processing 10,000
samples from a standard normal distribution. The network architecture is based
on a four-hidden layer network with 100, 100, 20, and one output neuron,
respectively. The top row shows the results by initializing the weights
according to $w \sim U\left(0,\sqrt{6/{(\operatorname{fan_{in}} +
\operatorname{fan_{out}})}}\right)$, the middle row by using a uniform
distribution with upper bound $b=\sqrt{\frac{3}{\operatorname{fan_{in}} +
\operatorname{fan_{out}}}}$, and the bottom row by using the initializing
function according as given in Equation~\ref{eq:initialize}. 
}
\label{fig:initialisation_activation} 
\end{figure}

While there seems to be little influence in the choice of hyperparameters,
architecture and learning rate, we found that the initialization of weights in
the monotonic NN $\phi^{-}_0$ plays a key role. This part of the \acronym{}
requires special attention as every weight is defined to be positive to
guarantee monotonicity. This can, e.g.,~be implemented in TensorFlow by using
the $\texttt{non\char`_neg}$ constraint function for every weight in every
layer. In addition to the constraint function, the initialization of the weights
should also be positive. In the following, we analyze three different
initializations. For the first initialization, we set the lower bound of the
Xavier initialization \cite{glorot2010understanding} to zero by sampling from $w
\sim U\left(0,\sqrt{6/{(\operatorname{fan}_{\text{in}} +
\operatorname{fan}_{\text{out}})}}\right)$, where
$\operatorname{fan}_{\text{in}}$ is the number of neurons in the previous layer
and $\operatorname{fan}_{\text{out}}$ the number or neurons in the current
layer. Using this naive approach, the variance after each layer increases, and
even a rather small neural network (e.g.,~with three hidden layers) will have
difficulty converging. To analyze the effect in more detail, we simulate data
from a standard normal distribution ($Y \sim N(0,1)$) and pass it through the
initialized $\phi^{-}_0$ network. The resulting distribution after each
activation function with the Xavier initialization and $\operatorname{tanh}$
activation is shown in the top row of
Figure~\ref{fig:initialisation_activation}. We see that after three hidden
layers, the activations in the network have saturated, making the training
extremely challenging.

The second initialization is based on the assumption that the expectation and
variance after each layer should remain constant. Using this assumption and by
using a uniform distribution with a zero lower bound it follows that the upper
bound $b$ should be initialized with
$b=\sqrt{\frac{3}{\operatorname{fan}_{\text{in}} +
\operatorname{fan}_{\text{out}}}}$. The middle row in Figure
\ref{fig:initialisation_activation} shows the activation distribution with this
initialization and the same input data. We see that this alternative
initialization scheme improves the saturation problem to some extent. However,
after the third layer, most of the activations are still saturated.

Further adapting the initialization, we empirically find that 
\begin{equation}
\label{eq:initialize} w \sim
U\left({0,\sqrt{\frac{9}{\operatorname{max}(\operatorname{fan_{\text{in}}},
\operatorname{fan_{\text{out}}})^2}}}\right) 
\end{equation}
results in only minor changes in the variance between different layers and
solves the convergence problems even for deeper architectures. 

\clearpage

\section{NUMERICAL EXPERIMENT DETAILS}

\subsection{General Parametrization} \label{app:generalparam}

In most cases where we use \acronym{} or neural basis functions, we specify 
\begin{itemize}
    \item $\phi_0^{-}$ by a monotonic neural network with two hidden layers,
    each with 10 hidden units, positive weight constraint and a tanh activation
    function. The hidden layers are followed by an output layer with 1 unit,
    also with positive weight constraint, but linear activation function to
    allow mapping into $\mathbb{R}$. 
    \item the neural basis functions $\rho_j$ by a multi-layer perception with a
    64-64-31-1 architecture, ReLU activations except for the last layer (which
    has no activation), and a bias in the penultimate layer only.
\end{itemize}

\subsection{UCI Benchmark Datasets}

Table~\ref{tab:further} provides an overview of UCI datasets used in our
benchmark.
\begin{table*}[!ht] \begin{center} \caption{Data set characteristics and
references.} \label{tab:further} \begin{tabular}{cccp{0.2cm}p{7cm}p{0.2cm}} 
\toprule
Data
set & \# Obs. & \# Feat. && Pre-processing \\ \midrule
Airfoil & 1503 & 5 && -  \\
Concrete   & 1030 & 8      &&  - \\ Diabetes  & 442 & 10      &&     -  \\
Energy  & 768 & 8          &&   -  \\ Fish & 908 & 6  && - \\ ForestF   & 517 &
12 &&  logp1 transformation for \texttt{area}; numerical representation for
\texttt{month} and \texttt{day} \\ Ltfsid & 182 & 4 && - \\ Naval & 11934 & 16
&& - \\ Real & 414 & 6 && Subtract minimum for \texttt{X1}; logp1 transformation
for \texttt{X2}; log transformation for \texttt{X3} -- \texttt{X6} as well as
for the outcome \\ Wine & 178 & 13 && - \\ Yacht  & 308 & 6           &&   - \\
\bottomrule \end{tabular} \end{center} \end{table*}

\subsection{Details of Individual Experiments}

\subsubsection{Mixture Modeling}

The Gaussian mixture (of experts) model implemented in a neural network as
suggested in \citet{ruegamer2022mixture} is parametrized by a mixture of two
normal distributions with the same additive predictor for both mean and standard
deviation parameter $\mu_\kappa, \sigma_\kappa, \kappa = 1,2$. The additive
predictor contains an intercept $\beta_{0,\mu,\kappa}$ and
$\beta_{0,\sigma,\kappa}$, respectively, and a thin-plate regression spline
$f_{\mu,\kappa}$,$f_{\sigma,\kappa}$, respectively, for the feature
$x_\text{time}$. 

\acronym{} is parametrized using a Bernstein polynomial of order 30 for
$\phi_0^{-}$ and a location shift using a neural basis function $\rho(x_{time})$
defined by a feed-forward neural network with two hidden layers, each with 64
units, ReLU activation and a bias term, and a final layer with 1 unit, linear
activation and no bias term.

\subsubsection{Ordinal Regression}

As in \cite{gal2015dropout}, we consider the subset of red wines and use the
same cross-validation folds. All features were standardized to the unit
interval. For each of the 20 splits, we fit a \texttt{tram::Polr}
\cite{pkg:tram} with linear covariate effects and a \acronym{} via
\texttt{deeptrafo::PolrNN} \cite{pkg:deeptrafo} with a neural basis function
architecture. The neural basis functions are specified via a fully connected
neural network with four ReLU layers with eight units each and a single unit
last layer with linear activation. The estimated partial effects for each
predictor were centered to have mean zero. The \acronym{} was trained for 200
epochs with the Adam optimizer, a learning rate of 0.001 and decay 0.0001.

\paragraph{Models and parameterizations} We have an ordered response $Y \in \{1,
\dots, K\}$ and covariates $X \in \RR^p$. The \acronym{} is parameterized with
the following base distribution $F$ and conditional inverse flow $\phi^-$:
\begin{align*}
F(\bcd) &\coloneqq \sigma(\bcd) = (1 + \exp(-\bcd))^{-1}, \\
\phi^-(y, x)  &\coloneqq \phi_0^-(y) + \sum_{j = 1}^p \rho_j(x_j), \\
\phi_0^-(y) &\coloneqq (\1(y = 1), \dots, \1(y = K))^\top\boldsymbol\theta\\
&\mbox{ s.t. } \theta_1 < \theta_2 < \dots < \theta_K \coloneqq +\infty,
\end{align*}
where $\rho_j : [0, 1] \to \mathbb{R}, \quad j = 1, \dots, p$ are
feed-forward neural networks as described above. 

The POLR model is the same as above with the specialization that $\rho_j(x_j)
\coloneqq x_j\beta_j$, $j = 1, \dots, p$ are linear functions.

\subsubsection{Survival Analysis}

We used the London Fire Brigade data set analyzed in \cite{taylor2017spatial}.
The data describes the response times of the respective fire brigade in relation
to spatiotemporal and economic features. We administratively censored excessive
response times ($T > 1000$) and model the censored event time with a spatial
effect (latitude and longitude), temporal effect (time of the day), and
categorical features for the property type and the district name. The predictors
$\rho_j$ are either linear effects for the categorical features or neural basis
functions for the time and spatial features with structure as described in
Section~\ref{app:generalparam}. More precisely, the temporal effect is modeled
with a univariate neural basis function with 64 and 12 units, each with ReLU
activation function. The spatial effect is modeled by a bivariate NAM with 64,
32, 32, and 10 units each and ReLU activation function. Using these effects, we
define the Cox proportional hazards model with smooth log cumulative baseline
hazards in our \acronym{} framework using the following parametrization:
\begin{align*}
    F : z &\mapsto 1 - \exp(-\exp(z)) \\
    \phi^-(y, x) &\coloneqq \boldsymbol{a}(y)^\top\boldsymbol{\theta} +
        \sum_{j=1}^p \rho_j(x_j) 
        , \\
    &\mbox{ s.t. } \theta_1 \leq \theta_2 \leq ... \leq \theta_{M+1},
\end{align*}
where $\Phi_{0,1}$ denotes the standard normal CDF and $\boldsymbol{a}$ a basis
of polynomials in Bernstein form of order $M$. are feed-forward neural networks.
We train the \acronym{} using the Adam optimizer with a learning rate of 0.001
in a batch size of 32 for 250 epochs. As a comparison, we fit a piece-wise
exponential additive model (PAM; \cite{bender2018}) with equivalent feature
effects using a thin-plate regression and tensor-product spline basis using
\texttt{pammtools} \cite{pammtools}. The evaluation criterion is the reweighted
integrated Brier Score introduced in \cite{sonabend2022scoring} which is a
proper scoring rule \cite{gneiting2007strictly}. Both learners are compared to
an uninformative model, the Kaplan-Meier estimator (KM;
\cite{kaplan1958nonparametric}) as a baseline.

\subsubsection{Generalized Additive Models}

We follow the data pre-processing of \cite{pho2023} and define both the NAM and
the GAM using a Poisson outcome distribution. Both additive predictors use
(neural-based) splines for date, population, temperature, and humidity. In
addition, a tensor-product spline is used for latitude and longitude. Non-linear
feature effects are defined by either 
\begin{itemize}
    \item[a)] univariate basis functions as described in
    Section~\ref{app:generalparam} to resemble univariate splines,
    \item[b)] two feed-forward neural networks as described in a) with 5 units
    in the last layer (instead of 1 unit) and then combined via a tensor-product
    followed by one last layer with 1 unit and no activation as well as no bias
    term.
\end{itemize}
The GAM uses thin-plate regression splines for univariate effects and a
tensor-product spline version for the bivariate spatial effect. The \acronym{'s}
inverse reference flow $\phi^{-}_0$ network is defined as a two-layer
non-negative tanh-network with 10 neurons each as well as positive weight
constraint. All models are trained for a maximum of 250 epochs, batch size of
128, early stopping based on a 10\% validation split with patience of 15 epochs,
and Adam with a learning rate of 0.001.

\subsubsection{Time Series Regression}

The electricity data set serves as a frequently chosen data set for
state-of-the-art forecasting challenges \cite[e.g.][]{Wang.2023, Wu.2023}. It
consists of records denoted in kilowatts at 15-minute intervals which we convert
to kilowatt-hours. In \acronym{}, the time lags enter linearly as location and
scale effects on the standard Gaussian base distribution. The \acronym{'s}
inverse reference flow $\phi^{-}_0$ network is defined as a three-layer
non-negative tanh-network with 20 neurons for the first two layers, 5 neurons in
the consecutive layer as well as positive weight constraint. The model for each
univariate time series trains for a maximum of 10,000 epochs with a batch size
of 256, and Adam as an optimizer with an initial learning rate of 0.0001. Early
stopping is based on the validation set described in the main text with a
patience of 10 epochs. For comparison, we employ the ARIMA model, a commonly
used benchmark \cite[e.g.][]{Siami.2018, Ruegamer.2023}. We specify
\texttt{auto.arima} such that the model complexity is found by a stepwise
forward search based on the biased-corrected version of Akaike’s Information
Criterion (AICc), with initial lag values of $p = 12$ and $p = 24$ for the
auto-regressive term, and $q = 0$ and $q = 3$ for the moving average term. We
set the maximum number of model search steps to 25. The final model, which
provides the log-scores obtained from the test set, is selected based on the
ARIMA model with the lowest AICc on the validation set. \acronym~runs 5.4 hours.
The auto.arima function in comparison has a runtime of 53 min.

\subsubsection{Benchmark Study} \label{app:benchmark}

DR is defined by a parametric normal distribution with mean $\mu = \sum_{j=1}^p
\rho_{\mu,j}$ and standard deviation $\sigma = \exp\left(\sum_{j=1}^p
\rho_{\sigma,j}\right),$ with additive predictor structure as explained in the
following paragraph. Training is done based on the negative log-likelihood. TMs
fit into our \acronym{} framework as a special case and are parameterized as
follows:
\begin{align*}
    F &\coloneqq \Phi_{0,1} \\ \phi^-(y, x) &\coloneqq
    \boldsymbol{a}(y)^\top\boldsymbol{\theta} + \sum_{j=1}^p \rho_j(x_j), \\
    &\mbox{ s.t. } \theta_1 \leq \theta_2 \leq ... \leq \theta_{M+1},
\end{align*}
where $\Phi_{0,1}$ denotes the standard normal CDF and $\boldsymbol{a}$ a basis
of polynomials in Bernstein form of order $M$. \acronym{} is defined by a
location and scale effect outlined in the following paragraph. For $\phi_0^{-}$
see Section~\ref{app:generalparam}. All models are trained using the Adam
optimizer with a maximum of 1000 epochs, early stopping with a patience of 50.

\paragraph{Predictor structure} \label{sss:pred}

\begin{itemize}
    \item \textbf{Structured} For comparisons of structured models, we use
    univariate thin-plate regression splines for DR and TMs for every feature
    and neural basis function splines as described in
    Section~\ref{app:generalparam}.
    \item \textbf{Deep} For deep model comparisons, we model the predictors of
    DR, TM and \acronym{} using four different multi-layer perceptron
    architectures:
\begin{itemize}
    \item Hidden(100,ReLU)-Dropout(0.1)-Hidden(1,Linear)
    \item
    Hidden(100,ReLU)-Dropout(0.1)-Hidden(100,ReLU)-Dropout(0.1)-Hidden(1,Linear)
    \item Hidden(20,ReLU)-Dropout(0.1)-Hidden(1,Linear)
    \item
    Hidden(20,ReLU)-Dropout(0.1)-Hidden(20,ReLU)-Dropout(0.1)-Hidden(1,Linear)
\end{itemize}
and for each method choose the best performing.
    \item \textbf{Semi-structured} For semi-structured comparisons, we use a
    combination of structured effects as outlined in the structured predictor
    section and one of the four deep neural networks as outlined in the deep
    predictor section.
\end{itemize}

\subsection{Further Results in the Benchmark Study}

In case \acronym~and comparison methods define a deep or semi-structured model,
we use a pre-defined set of four different deep architectures for all methods.
The results for these different model specifications are given in
Tables~\ref{tab:benchmark2}--\ref{tab:benchmark3}.

\begin{table}[!t]
\centering
\caption{\footnotesize Comparison results for different datasets (rows) and
structured methods using neural basis functions (columns) showing the mean
log-score (and standard deviation in brackets) based on a 10-fold
cross-validation. The best methods per dataset are highlighted in bold.}
\label{tab:benchmark2}
       \small
\begin{tabular}{lrrr}
\toprule
Dataset & DR & TM & DRIFT \\ \midrule
Airfoil & -3.5 (1.1) & \textbf{-3.1} (0.1) & -3.5 (0.7) \\
Concrete & -3.3 (0.1) & -3.4 (0.3) & \textbf{-3.2} (0.3) \\
Diabetes & -8.5 (2.5) & \textbf{-5.5} (0.3) & \textbf{-5.5} (0.2) \\
Energy & -2.9 (0.3) & -2.8 (0.4) & \textbf{-2.3} (0.1) \\
Fish & \textbf{-1.3} (0.1) & -1.4 (0.2) & \textbf{-1.3} (0.3) \\
ForestF & -2.0 (0.3) & -1.5 (0.2) & \textbf{-1.4} (0.4) \\
Ltfsid & -7.7 (7.5) & -6.3 (0.4) & \textbf{-4.8} (0.2) \\
Naval & \phantom{-}4.1 (1.0) & \phantom{-}3.6 (0.3) & \phantom{-}\textbf{5.1} (0.3) \\
Real & -1.4 (0.4) & \textbf{-0.8} (0.3) & -1.2 (1.1) \\
Wine & -0.2 (0.0) & \phantom{-}0.2 (0.2) & \phantom{-}\textbf{4.2} (2.4) \\
Yacht & -1.1 (0.2) & -2.5 (3.0) & \textbf{-0.8} (0.1) \\
\bottomrule
\end{tabular}

\end{table}

\begin{table}[!t]
\centering
\caption{\footnotesize Comparison results for different datasets (rows) and
\emph{deep predictor} methods (columns) showing the mean log-score (and standard
deviation in brackets) based on a 10-fold cross-validation. The best methods per
dataset are highlighted in bold.} \label{tab:benchmark1}
       \small
\begin{tabular}{lrrr}
\toprule
Dataset & DR & TM & DRIFT \\
\midrule
Airfoil & -4.4 (0.2) & \textbf{-3.1} (0.1) & -3.2 (0.3) \\
Concrete & -3.7 (0.1) & -3.4 (0.2) & \textbf{-3.2} (0.3) \\
Diabetes & -8.3 (1.2) & -5.7 (0.3) & \textbf{-5.2} (0.2) \\
Energy & -2.9 (0.1) & -3.1 (0.2) & \textbf{-2.5} (0.3) \\
Fish & \textbf{-1.3} (0.1) & -1.4 (0.2) & \textbf{-1.3} (0.3) \\
ForestF & -2.1 (0.5) & -1.6 (0.2) & \textbf{-1.2} (0.2) \\
Ltfsid & -5.7 (0.4) & -6.3 (0.4) & \textbf{-4.7} (0.1) \\
Naval & \phantom{-}4.7 (0.1) & \phantom{-}3.9 (0.2) & \phantom{-}\textbf{5.0} (0.2) \\
Real & -1.5 (0.6) & -1.2 (0.5) & \textbf{-0.9} (0.7) \\
Wine & -0.2 (0.0) & \phantom{-}0.7 (0.2) & \phantom{-}\textbf{4.2} (2.4) \\
Yacht & -1.6 (0.1) & -2.1 (1.6) & \textbf{-0.8} (0.1) \\
\bottomrule
\end{tabular}
\end{table}

\subsection{Computational Environment}

All real-world examples were conducted on conventional laptops and without GPU
support. 

\end{document}